\documentclass[letterpaper]{article} 
\usepackage{aaai24}  
\usepackage{times}  
\usepackage{helvet}  
\usepackage{courier}  
\usepackage[hyphens]{url}  
\usepackage{graphicx} 
\usepackage{natbib}  
\usepackage{caption} 
\usepackage{algorithm}
\usepackage{algorithmic}

\usepackage{amsmath}
\usepackage{amsthm}
\usepackage{amssymb}

\usepackage{multirow}
\usepackage{booktabs}

\usepackage{xcolor}

\usepackage{newfloat}
\usepackage{listings}
\DeclareCaptionStyle{ruled}{labelfont=normalfont,labelsep=colon,strut=off} 
\floatstyle{ruled}
\newfloat{listing}{tb}{lst}{}
\floatname{listing}{Listing}
\title{SGNet: Structure Guided Network via Gradient-Frequency Awareness \\ for Depth Map Super-Resolution}

\author{
    Zhengxue Wang, Zhiqiang Yan\thanks{Corresponding authors}, Jian Yang\footnotemark[1]
}
\affiliations{
    PCA Lab, Nanjing University of Science and Technology, China\\
    \{zxwang,yanzq,csjyang\}@njust.edu.cn
}

\usepackage{bibentry}

\begin{document}

\maketitle

\begin{abstract}
Depth super-resolution (DSR) aims to restore high-resolution (HR) depth from low-resolution (LR) one, where RGB image is often used to promote this task. Recent image guided DSR approaches mainly focus on spatial domain to rebuild depth structure. However, since the structure of LR depth is usually blurry, only considering spatial domain is not very sufficient to acquire satisfactory results. In this paper, we propose \emph{structure guided network} (SGNet), a method that pays more attention to gradient and frequency domains, both of which have the inherent ability to capture high-frequency structure. Specifically, we first introduce the \emph{gradient calibration module} (GCM), which employs the accurate gradient prior of RGB to sharpen the LR depth structure. Then we present the \emph{Frequency Awareness Module} (FAM) that recursively conducts multiple \emph{spectrum differencing blocks} (SDB), each of which propagates the precise high-frequency components of RGB into the LR depth. Extensive experimental results on both real and synthetic datasets demonstrate the superiority of our SGNet, reaching the state-of-the-art (see Fig.~\ref{fig:zzt}). Codes and pre-trained models are available at \url{https://github.com/yanzq95/SGNet}. 

\end{abstract}

\section{Introduction}
\begin{figure*}[t]
  \centering
    \includegraphics[width=0.93\linewidth]{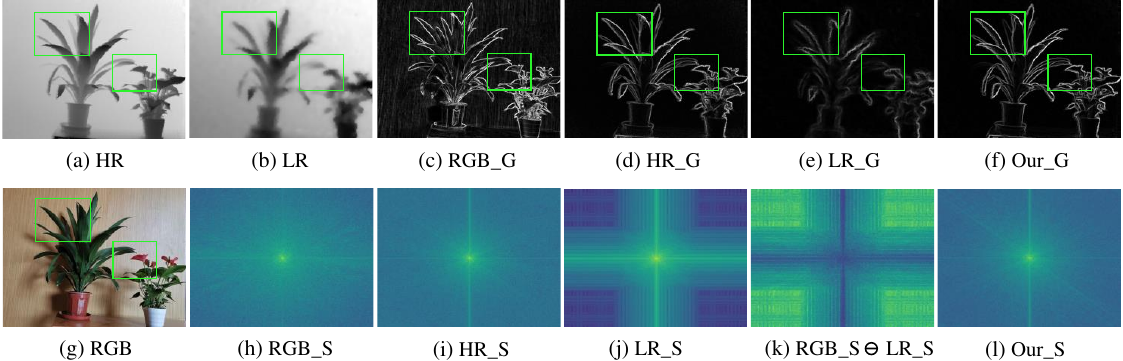}
    \caption{Visualizations of (c)-(f) \textbf{G}radient features and (h)-(l) \textbf{S}pectrum features, where $\ominus $ refers to subtraction.}
    \label{fig:freGrad}
\end{figure*}

Image guided DSR has been widely applied in various fields, such as 3D reconstruction~\cite{yuan2023recurrent}, virtual reality~\cite{bonetti2018augmented}, and augmented reality~\cite{xiong2021augmented}. However, the blurry structure of LR depth caused by complex imaging environment still impedes their performance. For example, Fig.~\ref{fig:freGrad} (LR) shows that, the LR depth contain rich low-frequency content but are severely deficient in clear high-frequency structure. Recently, many DSR approaches~\cite{yuan2023recurrent,shi2022symmetric} are proposed to tackle this issue. However, most of them focus only on the spatial domain for recovery, which is not very sufficient to obtain desired results.

\begin{figure}[t]
 \centering
 \includegraphics[width=0.95\columnwidth]{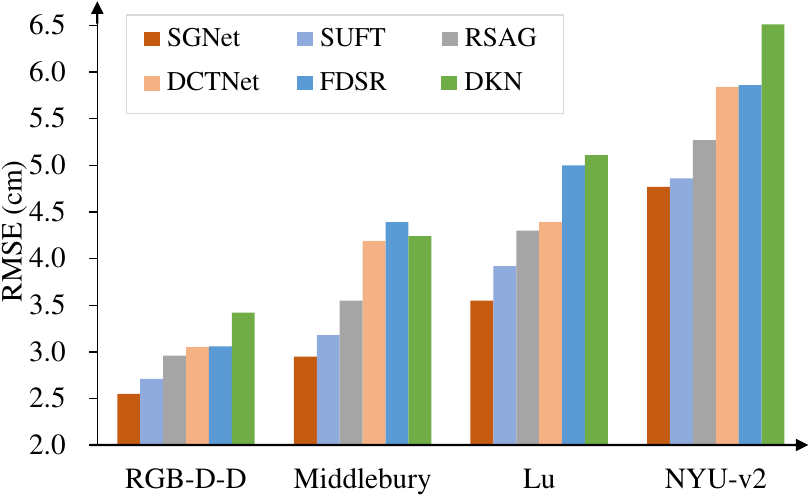}\\
 \caption{RMSE comparison between our SGNet and existing state-of-the-art methods on four benchmarks ( $\times 16$ ).}
 \label{fig:zzt}
\end{figure}
 
\emph{For one thing}, from (c)-(f) of Fig.~\ref{fig:freGrad} we discover that, the gradient features of RGB and HR contain highly discriminative object structure. Besides, although the degraded LR is terribly blurry, the gradient feature can still delineate its structure clearly. \emph{For another thing}, from (h)-(j) of Fig.~\ref{fig:freGrad} we find that, the spectrum features of RGB and HR reserve not only low-frequency content (central area) but also high-frequency structure (corner area). In contrast, the spectrum feature of LR lacks a large number of high-frequency components. These evidences indicate that the gradient and spectrum information can accurately depict the distribution of high-frequency structure. Consequently, \emph{motivated by these two observations, in this paper we pay more attention to gradient and frequency domains to take advantage of their inherent properties for clear structure recovery}.

\textbf{Gradient domain.} We design the \emph{gradient calibration module} (GCM) to leverage the powerful structure representation capability of gradient feature. Specifically, RGB and LR are first mapped into gradient domain~\cite{ma2020structure}. Then the accurate RGB gradient prior is employed to calibrate the blurry structure of LR. Besides, we introduce a gradient-aware loss to further sharpen the structure via narrowing the distance between the intermediate feature of GCM and that of HR in gradient domain.

\textbf{Frequency domain.} We present the \emph{Frequency Awareness Module} (FAM), which recursively conducts multiple \emph{spectrum differencing blocks} (SDB) to propagate the precise high-frequency components~\cite{yan2022rignet} of RGB. Concretely, SDB first maps RGB and LR into the same frequency domain. To explicitly compensate for the absent high-frequency components, \emph{i.e.}, the blank corner area of Fig.~\ref{fig:freGrad}(j), SDB next employs subtraction between the spectrum feature of RGB and that of LR in Fig.~\ref{fig:freGrad}(k), which is then merged with the spectrum feature of LR to enhance the structure. Besides, we initiate a frequency-aware loss to further strengthen the response of FAM in frequency space. 

Owing to the ingenious designs of GCM and FAM, (f) and (l) of Fig.~\ref{fig:freGrad} show that our approach can obtain very sharp and highlighted structure in gradient and frequency domains, respectively. As a result in Fig.~\ref{fig:zzt}, our SGNet surpasses the five state-of-the-art methods by \textbf{16\%} (RGB-D-D), \textbf{24\%} (Middlebury), \textbf{21\%} (Lu) and \textbf{15\%} (NYU-v2) in average. In summary, our contributions are as follows:

\begin{itemize}
\item Apart from the spatial domain, we introduce a novel perspective that exploits the gradient and frequency domains for the structure enhancement of DSR task. 

\item We propose SGNet that consists of novel GCM and FAM, where GCM leverages the gradient prior to adaptively calibrate and sharpen LR structure, whilst FAM employs recursive SDB to propagate the high-frequency components into LR for clear structure recovery. 

\item SGNet achieves significantly superior performance on both real-world and synthetic datasets. Codes and pre-trained models are released for peer research.  
\end{itemize}

\begin{figure*}[t]
  \centering
    \includegraphics[width=0.965\linewidth]{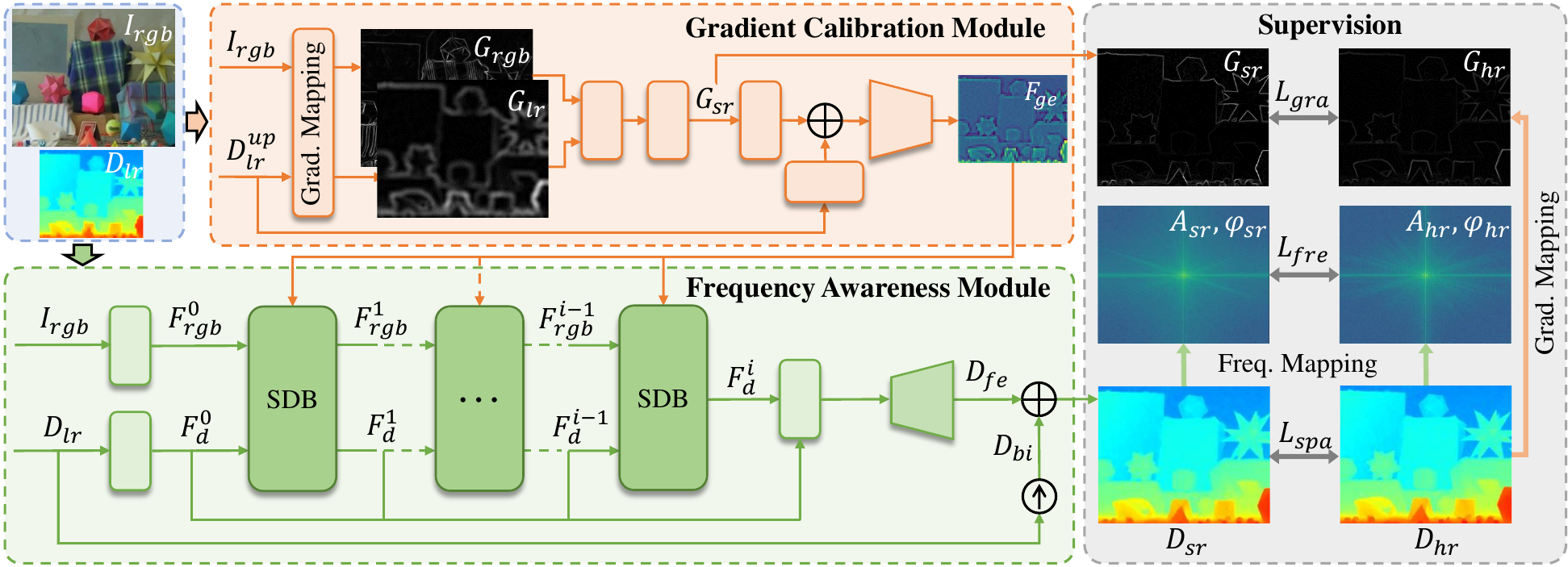}
    \caption{Overview of our Structure Guided Network (SGNet).  Given $I_{rgb}$ and $D_{lr}^{up}$ as input, the Gradient Calibration Module (GCM) first maps them into gradient domain, producing $F_{ge}$ with sharp depth structure.  Then, $I_{rgb}$, $D_{lr}$ and $F_{ge}$ are fed into the Frequency Awareness Module (FAM) to estimate frequency enhanced depth feature $D_{fe}$ via recursive Spectrum Differencing Blocks (SDB). $\uparrow$: bicubic up-sample. Grad. Mapping: Gradient Mapping. Freq. Mapping: Frequency Mapping.
    }
    \label{fig:SGNet}
\end{figure*} 

\section{Related Work}
\subsection{Depth Map Super-Resolution}
Benefiting from the rich structure of RGB image, guided DSR \cite{song2020channel,yang2022codon,zhong2021high} have attracted broad attention. For example, \cite{shi2022symmetric} introduce a symmetric uncertainty method to select RGB information that is effective for HR depth recovery whilst skipping harmful texture. \cite{kim2021deformable} design joint image filtering to adaptively output the neighbors and their weights for each pixel. \cite{deng2020deep} propose a multi-modal convolutional sparse coding to automatically split common and private features among different modalities. Similarly, \cite{zhao2022discrete} build a discrete cosine network that extracts both shared and specific multi-modal information through a semi-decoupled feature extraction module. Besides, some methods present multi-task learning frameworks to leverage complementary knowledge. For instance, \cite{yan2022learning} introduce an auxiliary depth completion branch to propagate the correlation of dense depth into the DSR branch. \cite{tang2021bridgenet} transmit RGB into a space that is close the depth space via depth estimation, thus facilitating the RGB-D fusion for DSR. Furthermore, \cite{sun2021learning} develop cross-task knowledge distillation to exchange correlation between DSR and depth estimation branches. Most recently, \cite{yuan2023recurrent} propose recursive structure attention to gradually estimate high-frequency structure. Meanwhile, \cite{yuan2023structure} design a structure flow-guided network to learn the edge-focused guidance feature for depth structure enhancement. In addition, graph regularization \cite{de2022learning} and anisotropic diffusion \cite{metzger2023guided} are applied to enhance the recovery of depth structure. Different from these approaches, most of which concentrate only on spatial domain, we pay more attention to gradient and frequency domains, employing the high-frequency components of RGB to guide depth structure. 

\subsection{Gradient and Frequency Learning}
Since the inherent characteristics of gradient and spectrum are quite helpful to represent structure, various related methods~\cite{sun2010gradient,lin2023catch} have been proposed. In gradient domain, \cite{qiao2023depth} employ low-cut filtering to extract gradient information and then fuse at multiple scales and stages for DSR. \cite{sun2010gradient} introduce a gradient field transformation to constrain the gradient fields of the HR image during the execution of SISR. \cite{ma2020structure} develop a structure-preserving method for single image super-resolution (SISR), which propagates gradient knowledge into the RGB branch. \cite{zhu2015modeling} build a gradient pattern dictionary to simulate deformable gradient of SISR. In frequency domain, \cite{zhou2022adaptively,zhou2022spatial} integrate spatial and spectrum features for multi-spectrum pan-sharpening. \cite{jiang2021focal} design a focal frequency loss to narrow the frequency domain gap between real image and generated image. \cite{mao2023intriguing} introduce a frequency selective network to adaptively learn kernel-level features for image deblurring. \cite{lin2023catch} carry out frequency-enhanced variational autoencoders to restore the high-frequency components lost during the image compression process. Inspired by these methods, we employ the gradient and spectrum of RGB to fully guide depth structure in both gradient and frequency domains.

\section{Method}

\subsection{Problem Formulation}
Given input LR depth $D_{lr} \in R^{h\times w\times 1}$ and HR RGB image $I_{rgb} \in R^{sh\times sw\times 3}$, guided DSR aims to predict HR depth $D_{sr} \in R^{sh\times sw\times 1}$ that is supervised by HR ground-truth depth $D_{hr} \in R^{sh\times sw\times 1}$. $h$, $w$, $s$ refer to the height, width, and scaling factor, respectively.

\subsection{Network Architecture}
As illustrated in Fig.~\ref{fig:SGNet}, the proposed SGNet is mainly composed of two modules, \emph{i.e.}, Gradient Calibration Module (GCM) and Frequency Awareness Module (FAM) that contains multiple Spectrum Differencing Blocks (SDB), aiming to recover more accurate depth structure in gradient and frequency domains, respectively. 

\emph{Firstly}, RGB $I_{rgb}$ and up-sampled depth $D_{lr}^{up}$ are fed into GCM, obtaining gradient representation $G_{rgb}$ and $G_{lr}$ by gradient mapping \cite{ma2020structure}. \emph{Secondly}, some residual groups~\cite{zhang2018image} and channel attention~\cite{woo2018cbam} are involved to calibrate the gradient of $G_{lr}$ via $G_{rgb}$, yielding $G_{sr}$ that is with the same resolution as $I_{rgb}$. \emph{Thirdly}, GCM outputs the guided feature $F_{ge}$ by down-sampling the sum of the gradient feature $G_{sr}$ and spatial feature $D_{lr}^{up}$. \emph{Fourthly}, $I_{rgb}$ and $D_{lr}$ are encoded to generate $F_{rgb}^{0}$ and $F_{d}^{0}$ respectively, both of which are then input into the first SDB of FAM together with $F_{ge}$. \emph{Fifthly}, FAM recursively conducts SDB to obtain $D_{fe}$. \emph{Finally}, the HR prediction $D_{sr}$ is produced by summing $D_{fe}$ and the bicubic interpolation of $D_{lr}$. During training process, our SGNet employs three loss functions of different domains, \emph{i.e.}, gradient-aware loss $\mathcal{L}_{gra}$, frequency-aware loss $\mathcal{L}_{fre}$ and spatial-aware loss $\mathcal{L}_{spa}$. Specifically, $\mathcal{L}_{gra}$ takes as input $G_{sr}$ and the gradient mapping of ground-truth $D_{hr}$, while $\mathcal{L}_{fre}$ inputs the corresponding mappings of $D_{sr}$ and $D_{hr}$.

\begin{figure}[t]
  \centering
     \includegraphics[width=0.98\linewidth]{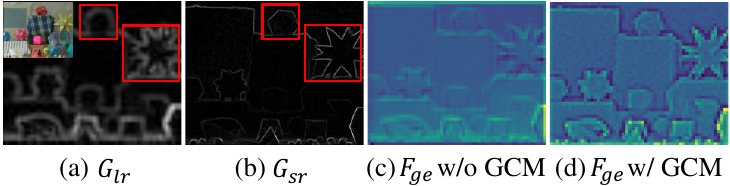}
   \caption{Visualization of (a)-(b) gradient features and (c)-(d) depth features on Middlebury dataset.}
   \label{fig:Grad_View} 
\end{figure}
\subsection{Gradient Calibration Module}
Our GCM is depicted in the orange part of Fig.~\ref{fig:SGNet}. We first employ a gradient mapping function~\cite{ma2020structure} $f_{gm}$ to transmit $I_{rgb}$ and $D_{lr}^{up}$ into gradient domain. Specifically, given input $Z$, the general definition of $f_{gm}$ is:
\begin{equation}
   	 f_{gm}  = \left \| \left ( Z_{x+1,y}-Z_{x-1,y}, Z_{x,y+1}-Z_{x,y-1}  \right )  \right \|_{2},
\end{equation}
where $Z_{x,y}$ refers to the pixel value at coordinates $\left ( x,y \right )$. Then the gradient features of $I_{rgb}$ and $D_{lr}^{up}$ are mapped as:
\begin{equation}
\begin{split}
    &G_{rgb}=f_{gm}\left( I_{rgb} \right), \\
    &G_{lr}=f_{gm}\left( D_{lr}^{up} \right).
\end{split}
\end{equation}

Next we conduct $f_g$ that is composed of two residual groups~\cite{zhang2018image} and a channel attention block~\cite{woo2018cbam}, to calibrate the gradient of $G_{lr}$, obtaining $G_{sr}\in R^{sh\times sw\times 1}$:
\begin{equation}
   	 G_{sr} = f_{g}\left ( G_{rgb},G_{lr}\right ).
\end{equation}

Supposing that the residual groups (orange and green rectangles in Fig.~\ref{fig:SGNet}) are denoted as $f_{r}$. Then, we fuse the gradient feature $G_{sr}$ and depth feature $D_{lr}^{up}$ to sharpen the depth structure. Finally, a down-sampling convolution $f_{ds}$ is deployed to decrease the resolution from $sh\times sw$ to $h\times w$, yielding the gradient-enhanced depth feature $F_{ge}$: 
\begin{equation}
   	 F_{ge} = f_{ds} \left ( f_{r}\left ( D_{lr}^{up}  \right ) + f_{r}\left ( G_{sr}  \right )   \right ).
\end{equation}

Fig.~\ref{fig:Grad_View} (b) and (d) show that our GCM successfully produce very clear gradient feature and sharp depth feature.

\subsection{Frequency Awareness Module}
 Our FAM is shown in the green part of Fig.~\ref{fig:SGNet}. The input $I_{rgb}$ and $D_{lr}$ are first mapped as:
\begin{equation}
\begin{split}
&F_{rgb}^{0} =f_{r} \left ( I_{rgb} \right ), \\
&F_{d}^{0} =f_{r} \left ( D_{lr} \right ).
\end{split}
\end{equation}
Next, given the gradient-enhanced feature $F_{ge}$, $F_{rgb}^{i}$ and $F_{d}^{i}$, FAM recursively conducts SDB to refine depth feature:
\begin{equation}
    F_{rgb}^{i},F_{d}^{i} = f_{s}^{i}  \left (F_{rgb}^{i-1},F_{d}^{i-1},F_{ge} \right ),\quad i\geq 1,
\end{equation}
where $f_{s}^{i}$ refers to \emph{i}th SDB. 
To leverage all of the history depth features, FAM concatenates $F_{d}^{0}, \cdots, F_{d}^{i}$. Then FAM encodes the combined feature via a residual group $f_r$ and an up-sampling convolutional layer $f_{up}$ to produce the frequency-enhanced feature $D_{fe}$. Finally, FAM outputs the predicted HR depth $D_{sr}$ by fusing $D_f$ with $D_{bi}$, \emph{i.e.}, the bicubic interpolation result of $D_{lr}$:
\begin{equation}
    \begin{split}
        &D_{sr} = D_{fe} + D_{bi}, \\
        &D_{fe}=f_{up}\left ( f_{r}\left ( \mathcal{C}_{i=1}^{n}{F_d^i}  \right ) \right ),
    \end{split}
\end{equation}
where $\mathcal{C}$ denotes concatenation. 

\begin{figure}[t]
  \centering
     \includegraphics[width=1\linewidth]{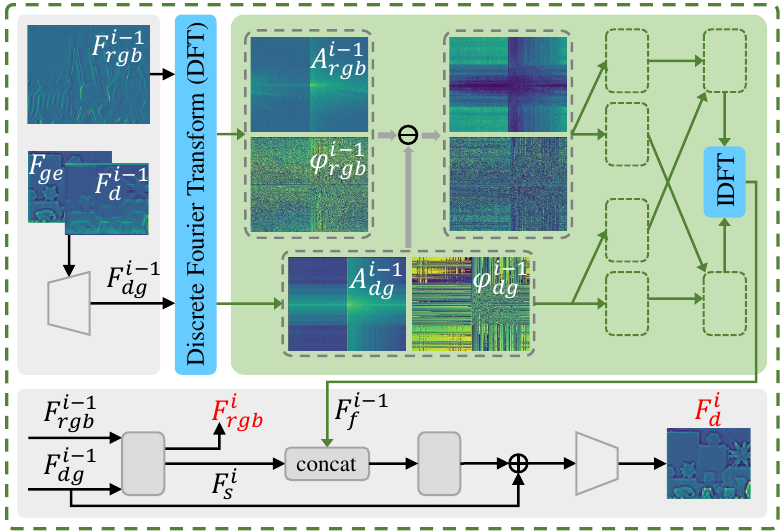}
   \caption{Spectrum differencing block (SDB). Green dashed box: $1\times 1$ convolution. Gray rectangular box: a $1\times 1$ convolution and an invertible neural network~\cite{zhou2022pan}.}
   \label{fig:SDM} 
\end{figure}

\subsubsection{Spectrum Differencing Block.}
Fig.~\ref{fig:SDM} shows that SDB first fuses and up-samples $F_{ge}$ and $F_{d}^{i-1}$ with a $3\times 3$ convolution $f_{c3}$ and $f_{up}$, producing HR $F_{dg}^{i-1}$:
\begin{equation}
F_{dg}^{i-1}=f_{up} \left ( f_{c3} \left ( \mathcal{C} \left ( F_{ge},F_{d}^{i-1}  \right ) \right )   \right ). 
\end{equation}
Next, the discrete fourier transform (DFT) $f_{df}$ is employed to map $F_{dg}^{i-1}$ and $F_{rgb}^{i-1}$ into the frequency domain, generating RGB spectrum $S_{dg}^{i-1}$ and depth spectrum $S_{rgb}^{i-1}$, both of which are decomposed into amplitude and phase:
\begin{equation}
\begin{split}
    & A_{dg}^{i-1},\varphi_{dg}^{i-1}  =f_{d}\left ( S_{dg}^{i-1} \right ) , \\
    & A_{rgb}^{i-1},\varphi_{rgb}^{i-1}  =f_{d}\left ( S_{rgb}^{i-1} \right ),
\end{split}
\end{equation}
where $S_{dg}^{i-1}=f_{df}\left ( F_{dg}^{i-1} \right )$ and $S_{rgb}^{i-1}=f_{df}\left ( F_{rgb}^{i-1} \right )$. $f_{d}$ denotes spectral decomposition function. $A_{dg}^{i-1}$ and $\varphi _{dg}^{i-1}$ refer to the amplitude and phase of $F_{dg}^{i-1}$, respectively, while  $A_{rgb}^{i-1}$ and $\varphi_{rgb}^{i-1}$ are the amplitude and phase of $F_{rgb}^{i-1}$.

AS depicted in the green part of Fig.~\ref{fig:SDM}, we first calculate the amplitude subtraction and phase subtraction between $S_{dg}^{i-1}$ and $S_{rgb}^{i-1}$ to produce $\left |A_{rgb}^{i-1}-A_{dg}^{i-1} \right |$ and $\left |\varphi_{rgb}^{i-1}-\varphi_{dg}^{i-1} \right |$, both of which are fed into multiple $1\times 1$ convolutional layers $f_{c}$ followed by activation function to learn high-frequency knowledge. Meanwhile, we also perform $f_{c}$ to extract the spectrum features of $A_{dg}^{i-1}$ and $\varphi_{dg}^{i-1}$. Then, the inverse discrete fourier transform (IDFT) $f_{idf}$ is employed to map the fused amplitude feature $A_{f}^{i-1}$ and phase feature $\varphi_{f}^{i-1}$ into spatial domain, producing $F_{f}^{i-1}$:
\begin{equation}
\begin{split}
    &F_{f}^{i-1} = f_{idf}\left (A_{f}^{i-1},  \varphi_{f}^{i-1} \right ), \\
    &A_{f}^{i-1}=f_{f}\left (f_{c}\left ( A_{dg}^{i-1} \right ),f_{c}\left ( \left |A_{rgb}^{i-1}-A_{dg}^{i-1} \right | \right ) \right ), \\
    &\varphi_{f}^{i-1}=f_{f}\left ( f_{c}\left ( \varphi_{dg}^{i-1} \right ),f_{c}\left ( \left |\varphi_{rgb}^{i-1}-\varphi_{dg}^{i-1} \right |  \right ) \right ),
\end{split}
\end{equation}
where $f_{f}$ consists of a $1\times 1$ convolution and concatenation. 

\begin{figure}[t]
  \centering
     \includegraphics[width=0.98\linewidth]{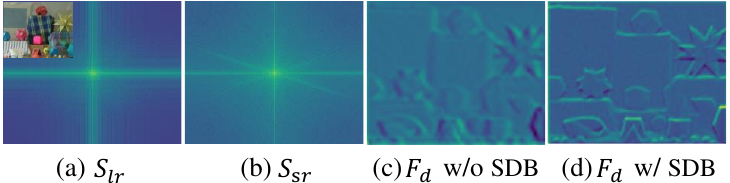}
   \caption{Visualization of (a)-(b) spectrum features and (c)-(d) depth features on Middlebury dataset. $S_{lr}$: LR spectrum. $S_{sr}$: predicted spectrum.  }
   \label{fig:Fre_View} 
\end{figure}

Finally, to strengthen the correlation between spatial domain and frequency domain, we employ the invertible neural network $f_{i} $~\cite{zhou2022pan} and down-sampling $f_{ds}$ to fuse the spatial domain feature $F_{s}^{i}$ and $F_{f}$:

\begin{equation}
    F_{d}^{i} = f_{ds} \left ( F_{dg}^{i-1} + f_{i} \left ( \mathcal{C} \left ( F_{s}^{i},F_{f}^{i-1} \right ) \right ) \right ),
\end{equation}
where $F_{rgb}^{i},F_{s}^{i}=f_{i}\left ( \mathcal{C} \left ( F_{rgb}^{i-1}, F_{dg}^{i-1}  \right )\right )$. $F_{d}^{i}$ and $F_{rgb}^{i}$ refer to the output depth feature and RGB feature of \emph{i}th SDB, respectively. As shown in Fig.~\ref{fig:Fre_View} (b) and (d), our SDB succeeds recovering high-frequency components and clear structure. 

\begin{figure*}[t]
\centering
\includegraphics[width=0.995\linewidth]{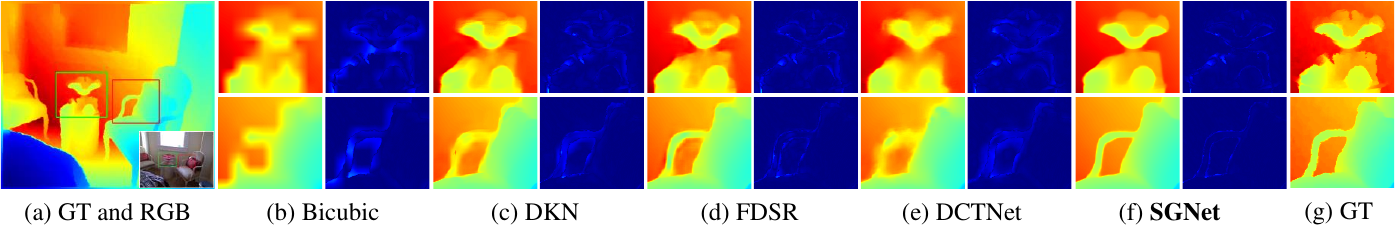}
\caption{Visual results and error maps on NYU-v2 dataset ($\times16$). Brighter color in error maps refers to larger error.}
\label{fig:NYU_X16}
\end{figure*}

\begin{table*}[t]
	\centering
	\resizebox{0.996\linewidth}{!}{
\begin{tabular}{c|cccccccccccccc}
\toprule[1.2pt] 
Scale&Bicubic &TGV &DJF & DMSG & GbFT  & DKN& FDSR &CTKT&DCTNet&AHMF & RSAG&SUFT&\textbf{SGNet}\\
\midrule
$\times4$ &8.16& 4.98 &3.54& 3.02    &3.35   & 1.62 & 1.61 &1.49  &1.59		&1.40	&1.23& \underline{1.12}	& \textbf{1.10}     	\\
$\times8$ &14.22& 11.23 &6.20 &2.99   &5.73     & 3.26 & 3.18 &2.73   &3.16	&2.89	&\underline{2.51} & \underline{2.51}	& \textbf{2.44}	\\
$\times16$&22.32&28.13&10.21 & 9.17 &9.01  & 6.51 & 5.86 &5.11   &5.84		&5.64	&5.27& \underline{4.86}	& \textbf{4.77}		\\
\bottomrule[1.2pt]
\end{tabular}}
\caption{Quantitative comparison with existing  state-of-the-art methods on NYU-v2 dataset.}\label{tab:NYU-v2}
\end{table*}

\begin{figure*}[t]
\centering
\includegraphics[width=0.995\linewidth]{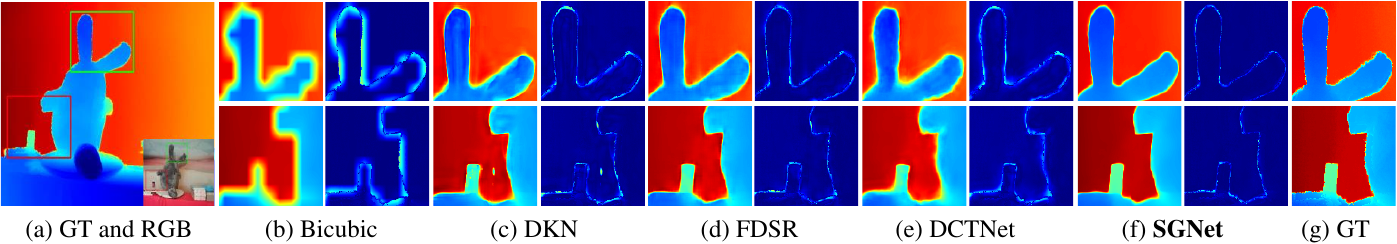}
\caption{Visual results and error maps on  RGB-D-D dataset ($\times16$).}
\label{fig:RGBDD_X16}
\end{figure*}

\begin{table*}[t]
\footnotesize
	\centering
	\resizebox{0.996\linewidth}{!}{
\begin{tabular}{c|ccccccccccccc}
\toprule[1.2pt] 
Scale&Bicubic &SDF &DJF & PAC & DJFR  & DKN& FDKN &FDSR&JIIF &DCTNet & RSAG&SUFT&\textbf{SGNet}\\
\midrule
$\times4$ &2.00&4.06& 3.41 &1.25& 3.35    &1.30  &1.18 &1.16 &1.17 &\textbf{1.08}  &1.14		& \underline{1.10}	& \underline{1.10}     	\\
$\times8$ &3.23&5.51& 5.57 &1.98 &5.57   &1.96     &1.91 &1.82 &1.79 &1.74   &1.75	& \underline{1.69}	& \textbf{1.64}	\\
$\times16$&5.16&7.39&8.15 &3.49 &7.99   &3.42  &3.41 & 3.06 &2.87 &3.05   &2.96		& \underline{2.71}	& \textbf{2.55}		\\
\bottomrule[1.2pt]
\end{tabular}}
\caption{Quantitative comparison with existing  state-of-the-art methods on RGB-D-D dataset.}\label{tab:RGB-D-D}
\end{table*}

\begin{table}[t]
	\centering
    \small
	\resizebox{0.93\linewidth}{!}{
\begin{tabular}{l|ccc|ccc}
\toprule[1.2pt]
 \multirow{2}{*}{Methods} & \multicolumn{3}{c|}{Middlebury } &  \multicolumn{3}{c}{Lu}\\
 \cmidrule{2-7}
&$\times4$ &$\times8$ &$\times16$ &$\times4$ &$\times8$ &$\times16$\\
\midrule
 DJF&1.68&3.24&5.62&1.65&3.96&6.75\\
 DJFR&1.32&3.19&5.57&1.15&3.57&6.77\\
 FDKN&\underline{1.08}&2.17&4.50&\underline{0.82}&2.10&5.05\\
 DKN &1.23&2.12&4.24&0.96 &2.16&5.11  \\
 FDSR&1.13 &2.08&4.39&1.29 &2.19&5.00\\
DCTNet&1.10 &2.05&4.19 &0.88&1.85&4.39\\
 RSAG&1.13&\underline{1.74}&3.55&\textbf{0.79}&\underline{1.67}&4.30\\
SUFT &\textbf{1.07}&1.75&\underline{3.18}&1.10 &1.74&\underline{3.92}\\
 \midrule
\textbf{SGNet}&1.15 &\textbf{1.64}&\textbf{2.95} &1.03&\textbf{1.61}&\textbf{3.55}\\
\bottomrule[1.2pt]
\end{tabular}}
\caption{Quantitative comparison with existing  state-of-the-art methods on Middlebury dataset and Lu dataset.}\label{tab:Lu_Midd}
\end{table}

\subsection{Loss Function}

Given $N$ RGB-D pairs, a spatial-aware loss $\mathcal{L}_{spa}$ is used:
\begin{equation}
   	\mathcal{L}_{spa}= \frac{1}N{\sum_{i=1}^{N} \left \| D_{sr}^{i} -D_{hr}^{i} \right \|_{1}}.
\end{equation} 
Next we utilize a gradient-aware loss $\mathcal{L}_{gra}$ to facilitate the calibration of LR gradient information:
\begin{equation}
   	\mathcal{L}_{gra}= \frac{1}N{\sum_{i=1}^{N} \left \| G_{sr}^{i} - G_{hr}^{i} \right \|_{1}},
\end{equation}
where $G_{hr}^{i}$ denote the ground-truth gradient. 

Then we introduce a frequency-aware loss $\mathcal{L}_{fre}$ to learn the HR spectrum, which consists of an amplitude loss $\mathcal{L}_{amp}$ and a phase loss $\mathcal{L}_{pha}$:
\begin{equation}
\begin{split}
&\mathcal{L}_{fre}= \lambda _{1} \mathcal{L}_{amp}+\lambda _{2} L_{pha}, \\
    &\mathcal{L}_{amp}= \frac{1}N{\sum_{i=1}^{N} \left \|  A_{sr}^{i} -  A_{hr}^{i} \right \|_{1}}, \\
   &\mathcal{L}_{pha }= \frac{1}N{\sum_{i=1}^{N} \left \|  \varphi _{sr}^{i} -  \varphi _{hr}^{i} \right \|_{1}},
\end{split}
\end{equation}
where $A_{sr}$ and $\varphi _{sr}$ severally refer to the amplitude and phase of the HR depth prediction, while $A_{hr}$ and $\varphi _{hr}$ correspond to those of the ground-truth depth. $\lambda _{1}$ and $\lambda _{2}$ are hyper-parameters.

Finally, the total training loss is defined as:
\begin{equation}
   	\mathcal{L}_{total}= \mathcal{L}_{spa} + \gamma _{1} \mathcal{L}_{gra} + \gamma _{2} \mathcal{L}_{fre},
\end{equation}
where $ \gamma _{1}$, $ \gamma _{2}$ are hyper-parameters.

\section{Experiments}

\subsection{Experimental Settings}
\subsubsection{Datasets.}

We conduct experiments on both synthetic NYU-v2~\cite{silberman2012indoor}, Middlebury~\cite{hirschmuller2007evaluation,scharstein2007learning}, Lu~\cite{lu2014depth}, and real-world RGB-D-D~\cite{he2021towards} datasets. Following previous works \cite{sun2021learning,yuan2023recurrent,zhao2023spherical}, on NYU-v2 dataset, the training set contains 1000 RGB-D pairs, while the test set consists of 449 pairs. Besides, the pre-trained model on NYU-v2 is also tested on Middlebury (30 pairs), Lu (6 pairs), and RGB-D-D (405 pairs). In these synthetic scenarios, the LR depth input is produced by bicubic down-sampling from the HR depth ground-truth. To validate the generalization of our method in real-world environment, we implement our SGNet on the real-world RGB-D-D dataset, including 2,215 RGB-D pairs for training and 405 for testing, where the LR depth is obtained via the ToF camera of Huawei P30 Pro.

\subsubsection{Metrics and Implementation Details.}
Following previous methods \cite{kim2021deformable,zhao2022discrete}, the \textbf{root mean square error (RMSE) in centimeter} is employed as the evaluation metric. During the training, we randomly crop the RGB image and HR depth into $256\times 256$. Adam optimizer \cite{Kingma2014Adam} with an initial learning rate of $1\times 10^{-4}$ is used to train SGNet with a single TITAN RTX GPU. The hyper-parameters are set as $\lambda _{1}=\lambda _{2}=0.5 $, $\gamma  _{1}=0.001$ and $\gamma  _{2}=0.002$. 

\subsection{Comparison with the state-of-the-art}
Tabs.~\ref{tab:NYU-v2}-\ref{tab:RGBDD_Real} compare SGNet with state-of-the-art methods on $\times 4$, $\times 8$ and $\times 16$ DSR, including TGV~\cite{ferstl2013image}, SDF~\cite{ham2017robust}, DJF~\cite{li2016deep}, DJFR~\cite{li2019joint}, PAC~\cite{su2019pixel}, DMSG~\cite{hui2016depth}, GbFT~\cite{albahar2019guided}, DKN~\cite{kim2021deformable}, FDKN~\cite{kim2021deformable}, FDSR~\cite{he2021towards}, JIIF~\cite{tang2021joint}, CTKT~\cite{sun2021learning}, AHMF~\cite{zhong2021high}, DCTNet~\cite{zhao2022discrete},  SUFT~\cite{shi2022symmetric} and RSAG~\cite{yuan2023recurrent}.

\subsubsection{Quantitative Comparison.}
Overall, Tabs.~\ref{tab:NYU-v2}-\ref{tab:RGBDD_Real} report that our SGNet achieves state-of-the-art performance on both the synthetic (Tabs.~\ref{tab:NYU-v2}-\ref{tab:Lu_Midd}) and real-world (Tab.~\ref{tab:RGBDD_Real}) datasets.
Specifically, From Tabs.~\ref{tab:NYU-v2} and~\ref{tab:RGB-D-D} we observe that SGNet is superior to the most of other methods on NYU-v2 and RGB-D-D datasets. For example, compared to the second best methods, our SGNet decreases the RMSE by $0.09cm$ (NYU-v2) and $0.16cm$ (RGB-D-D) on $\times 16$ DSR. Besides, Tab.~\ref{tab:Lu_Midd} verifies the generalization ability of our SGNet on Middlebury and Lu datasets. We discover that our SGNet is significantly superior to others on $\times 8$ and $\times 16$ DSR and obtains competitive result on $\times 4$ DSR. When comparing to the suboptimal methods on $\times 16$ DSR, the RMSE of our SGNet is $0.23cm$ (Middlebury) and $0.37cm$ (Lu) lower. 

Tab.~\ref{tab:RGBDD_Real} lists the comparison on the real-world RGB-D-D dataset. Following previous methods \cite{zhao2022discrete,yuan2023recurrent}, we first test the pre-trained models of NYU-v2 on RGB-D-D (without *), then retrain and test FDSR, DCTNet, SUFT and SGNet on RGB-D-D (with *). We find that both our SGNet and SGNet* achieve the best performance. For example, SGNet* surpasses the second best SUFT* by $0.09cm$ in RMSE. In short, these evidences confirm that SGNet contributes to better performance and generalization.

\subsubsection{Visual Comparison.}
\begin{figure}[t]
  \centering
     \includegraphics[width=0.99\linewidth]{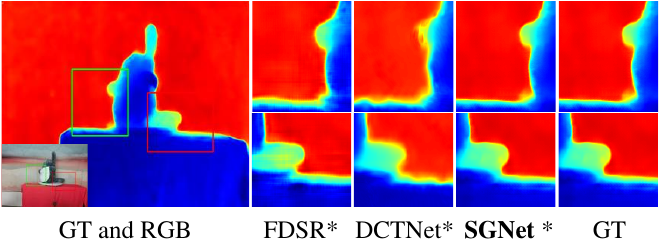}
   \caption{Visual comparison on the real-world RGB-D-D.}
   \label{fig:RGBDD_Real} 
\end{figure}

Figs.~\ref{fig:NYU_X16}-\ref{fig:RGBDD_Real} present visual comparison results. Obviously, our method can restore more precise depth predictions with clearer and sharper structure. For example, the edges of chair and doll in ($\times 16$) in Figs.~\ref{fig:NYU_X16} and~\ref{fig:RGBDD_X16} are more discriminative than others, while the errors maps shows the higher accuracy. Although previous methods that mainly focus on spatial domain are able to recover the most of depth information, the detailed structure is still difficult to predict. In contrast, our SGNet pays more attention to both gradient and frequency domains to leverage their inherent advantages of modeling structure. 

Furthermore, the DSR task is more challenging in real-world scenarios because the LR depth is often blurry and distorted. Fig.~\ref{fig:RGBDD_Real} demonstrates the visual comparison on the real-world RGB-D-D dataset. Compared to the state-of-the-art FDSR* and DCTNet*, our SGNet* can obtain more accurate HR depth with clear structure that is very close to the ground-truth depth. These visual results validate that our method can effectively enhance depth structure. 

\begin{table}[t]
\huge
	\centering
	\resizebox{0.98\linewidth}{!}{
\begin{tabular}{lcc|lcc}
\toprule[1.2pt] 
Methods & Train    & RMSE   & Methods & Train    & RMSE \\
\midrule
DJF     & NYU-v2   & 7.90   & DCTNet  & NYU-v2   &7.37 \\
DJFR    & NYU-v2   & 8.01   & SUFT    & NYU-v2   & \textbf{7.22} \\
FDKN    & NYU-v2   & 7.50   & FDSR*   & RGB-D-D   &5.49 \\
DKN     & NYU-v2   & 7.38   & DCTNet* & RGB-D-D   &5.43 \\
FDSR    & NYU-v2   & 7.50   & SUFT*   & RGB-D-D   &\underline{5.41} \\
 \midrule
\textbf{SGNet}    & NYU-v2  & \textbf{7.22}    &\textbf{SGNet*}  & RGB-D-D &\textbf{5.32}   \\
\bottomrule[1.2pt]
\end{tabular}}
\caption{Quantitative comparison on real-world RGB-D-D.}\label{tab:RGBDD_Real}
\end{table}

\subsection{Ablation Study}
\subsubsection{GCM and FAM.}
\begin{figure}[t]
 \centering
 \includegraphics[width=1\columnwidth]{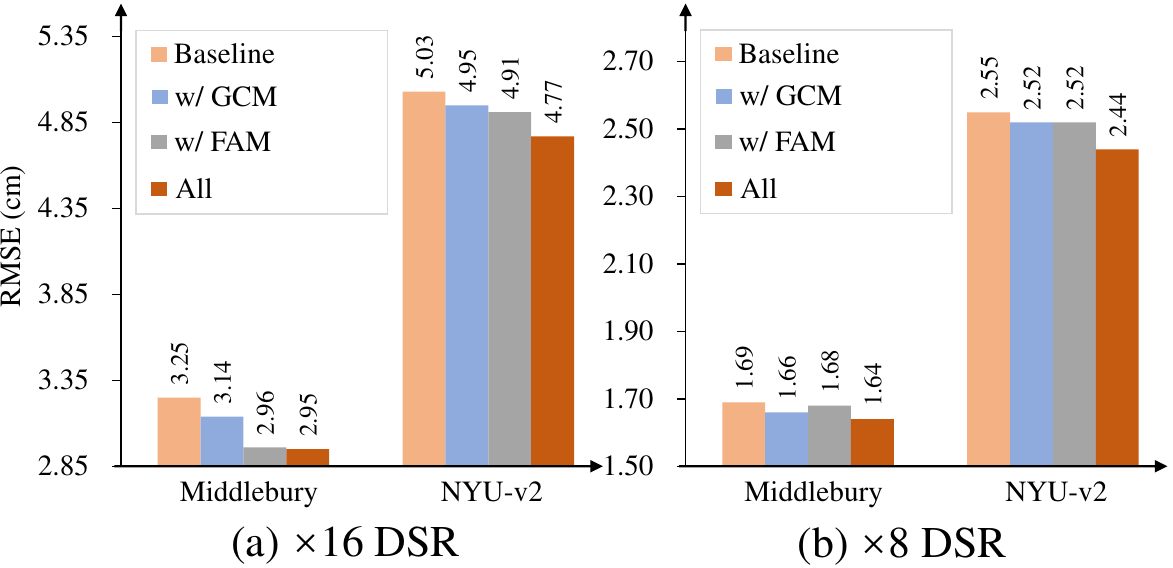}\\
 \caption{Ablation study of GCM and FAM of SGNet.}
 \label{fig:zzt2}
\end{figure}

\begin{figure}[t]
  \centering
     \includegraphics[width=0.99\linewidth]{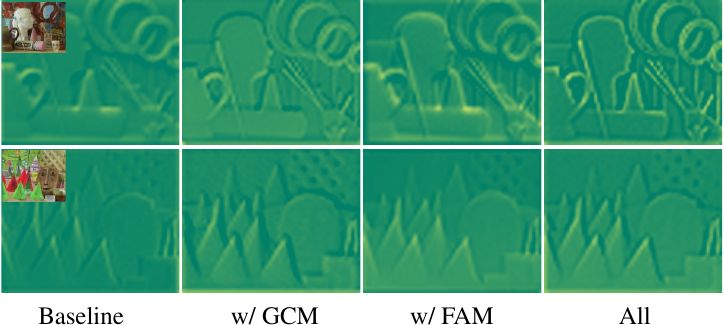}
   \caption{Visual comparison of intermediate depth features on Middlebury dataset ($\times 8$ case).}
   \label{fig:Ablation} 
\end{figure}

Figs.~\ref{fig:zzt2} and~\ref{fig:Ablation} show the ablation study of GCM and FAM. For the baseline, we first remove GCM completely. Then in SDB (Fig.~\ref{fig:SDM}) of FAM, the frequency operation is removed, only retaining the bottom gray part. 

As shown in Fig.~\ref{fig:zzt2}, both GCM and FAM can reduce RMSE by propagating the high-frequency components of RGB into depth in the gradient and frequency domains. When combining GCM and FAM, SGNet achieves the best performance. For example, compared to the $\times 16$ baseline, GCM is $0.11cm$ superior on Middlebury and $0.08cm$ on NYU-v2. FAM also reduces the error by $0.29cm$ and $0.12cm$, respectively. Finally, our SGNet surpasses the baseline by $0.3cm$ on Middlebury and $0.26cm$ on NYU-v2. 

Tab.~\ref{Params} lists the complexity comparison. It is observed that GCM \& FAM are lightweight, \textit{i.e.}, 1.68M \& 0.39M cost in Parameters, 1.30G \& 1.53G cost in Memory, and 235.0G \& 42.6G cost in FLOPs, respectively. When GCM \& FAM are combined to, the performance is significantly improved with acceptable sacrifice in complexity. 

In addition, Fig.~\ref{fig:Ablation} illustrates the visual results of intermediate depth features. We discover that both GCM and FAM are able to generate clearer depth structure than the baseline. Furthermore, when GCM and FAM are deployed together, our SGNet can produce much sharper structure.  

These numerical and visual evidences indicate that, by exploiting the structure representation in gradient and frequency domains, our SGNet can significantly enhance depth structure and thus improve the performance.

\begin{table}[t]
	\centering
	    \resizebox{1\linewidth}{!}{ 
		\begin{tabular}{l|cccc}
			\toprule[1.2pt] 
			Methods   	&Params (M) &FLOPs (G) 	&Memory (G)	&Time (ms)    	\\ 
			\midrule
			Baseline   	&37.18 $\left ( \pm   0.00 \right )$ 	&4346.3 $\left ( \pm   0.0 \right )$ 		&8.02 $\left ( \pm   0.00 \right )$    &63 $\left ( \pm   0 \right )$      \\ 
			+GCM        &38.86 $\left ( +  1.68 \right )$	    &4581.3 $\left ( +  235.0 \right )$		    &9.32 $\left ( +  1.30 \right )$       &69 $\left ( +  6 \right )$   \\
                +FAM        &37.57 $\left ( +  0.39 \right )$	    &4388.9 $\left ( +  42.6 \right )$		    &9.55 $\left ( +  1.53 \right )$   &67 $\left ( +  4 \right )$      \\
			All    &39.25 $\left ( +  2.07 \right )$ 	    &4623.9 $\left ( +  277.6 \right )$	        &10.77$\left ( +  2.75 \right )$	      &73 $\left ( + 10  \right )$	  \\
			\bottomrule [1.2pt] 
		\end{tabular}}
		\caption{Complexity on NYU ($\times 8$) tested by a 4090 GPU.}
		\label{Params}
\end{table}

\begin{figure}[t]
 \centering
 \includegraphics[width=0.98\columnwidth]{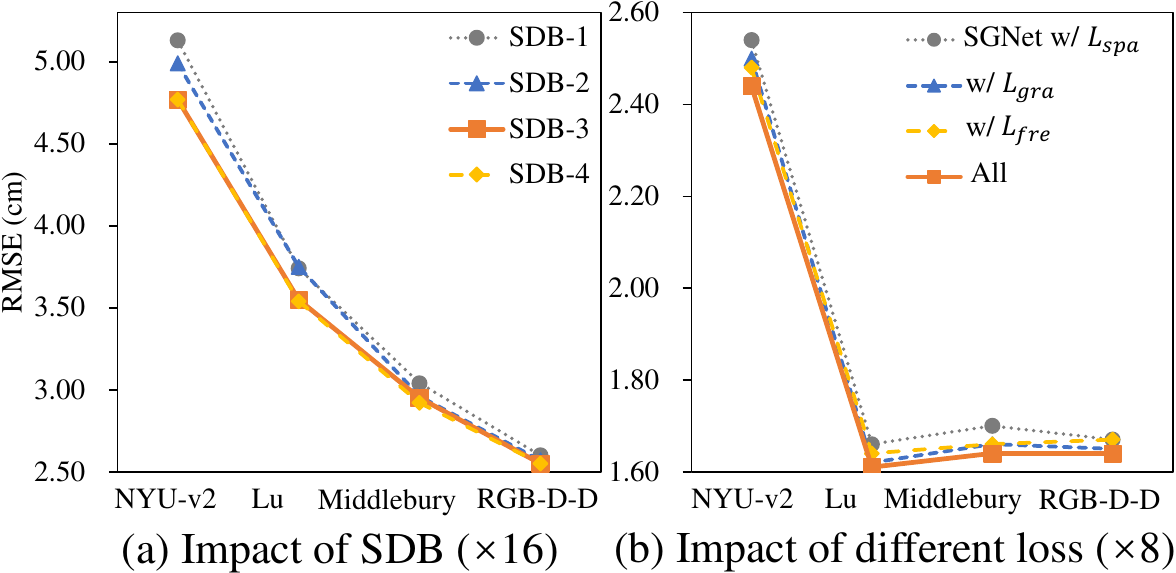}\\
 \caption{Ablation study of SGNet with (a) different numbers of SDB and (b) different loss functions.}
 \label{fig:zxt}
\end{figure}

\subsubsection{Different Recursion Numbers of SDB.}
Fig.~\ref{fig:zxt} (a) shows the ablation study of different numbers of SDB. The baseline is SGNet with GCM, FAM and all loss functions. It can be observed that the performance gradually improves as the number of SDB increases. When employing SDB-4, the errors decrease a little on Lu and Middlebury but maintain unchanged on NYU-v2 and RGB-D-D. Therefore, for better trade-off between the model complexity and accuracy, we select SDB-3 ( the orange solid line) as the default setting.

\subsubsection{Different Loss Functions.}
Fig.~\ref{fig:zxt} (b) shows the ablation study of different loss functions. The baseline is SGNet with GCM, FAM and $\mathcal{L}_{spa}$ only. We find that both the gradient-aware $\mathcal{L}_{gra}$ and frequency-aware $\mathcal{L}_{fre}$ contribute to performance improvement. When $\mathcal{L}_{gra}$ and $\mathcal{L}_{fre}$ are deployed together, the model achieves the best performance, \emph{i.e.}, averagely surpassing the basic SGNet by $0.06cm$ on NYU-v2, Lu, Middlebury and RGB-D-D datasets.

\section{Conclusion}
In this paper, we proposed SGNet, a novel DSR solution that payed more attention to gradient and frequency domains, employing the high-frequency components of RGB to enhance depth structure. For gradient domain, we designed the gradient calibration module to adaptively sharpen the blurry structure of LR depth via clear RGB gradient prior. For frequency domain, we developed the frequency awareness module, which recursively conducted multiple spectrum differencing blocks to propagate the high-frequency knowledge of RGB spectrum into the depth. Besides, we introduced the gradient-aware loss and frequency-aware loss to further narrow the structure distance of the prediction and target in both gradient and frequency domains. Extensive experiments demonstrated that our SGNet achieved state-of-the-art performance on four benchmark datasets.

\section{Acknowledgements}
The authors thank all reviewers for their instructive comments. This work was supported by the Postgraduate Research \& Practice Innovation Program of Jiangsu Province (KYCX23\_0471). Note that the PCA Lab is associated with, Key Lab of Intelligent Perception and Systems for High-Dimensional Information of Ministry of Education, and Jiangsu Key Lab of Image and Video Understanding for Social Security, School of Computer Science and Engineering, Nanjing University of Science and Technology.

\bibliography{aaai24}

\begin{thebibliography}{44}
\providecommand{\natexlab}[1]{#1}

\bibitem[{AlBahar and Huang(2019)}]{albahar2019guided}
AlBahar, B.; and Huang, J.-B. 2019.
\newblock Guided image-to-image translation with bi-directional feature transformation.
\newblock In \emph{ICCV}, 9016--9025.

\bibitem[{Bonetti, Warnaby, and Quinn(2018)}]{bonetti2018augmented}
Bonetti, F.; Warnaby, G.; and Quinn, L. 2018.
\newblock Augmented reality and virtual reality in physical and online retailing: A review, synthesis and research agenda.
\newblock \emph{Augmented reality and virtual reality: Empowering human, place and business}, 119--132.

\bibitem[{De~Lutio et~al.(2022)De~Lutio, Becker, D'Aronco, Russo, Wegner, and Schindler}]{de2022learning}
De~Lutio, R.; Becker, A.; D'Aronco, S.; Russo, S.; Wegner, J.~D.; and Schindler, K. 2022.
\newblock Learning graph regularisation for guided super-resolution.
\newblock In \emph{CVPR}, 1979--1988.

\bibitem[{Deng and Dragotti(2020)}]{deng2020deep}
Deng, X.; and Dragotti, P.~L. 2020.
\newblock Deep convolutional neural network for multi-modal image restoration and fusion.
\newblock \emph{IEEE transactions on pattern analysis and machine intelligence}, 43(10): 3333--3348.

\bibitem[{Ferstl et~al.(2013)Ferstl, Reinbacher, Ranftl, R{\"u}ther, and Bischof}]{ferstl2013image}
Ferstl, D.; Reinbacher, C.; Ranftl, R.; R{\"u}ther, M.; and Bischof, H. 2013.
\newblock Image guided depth upsampling using anisotropic total generalized variation.
\newblock In \emph{ICCV}, 993--1000.

\bibitem[{Ham, Cho, and Ponce(2017)}]{ham2017robust}
Ham, B.; Cho, M.; and Ponce, J. 2017.
\newblock Robust guided image filtering using nonconvex potentials.
\newblock \emph{IEEE transactions on pattern analysis and machine intelligence}, 40(1): 192--207.

\bibitem[{He et~al.(2021)He, Zhu, Li, Bai, Cong, Zhang, Lin, Liu, and Zhao}]{he2021towards}
He, L.; Zhu, H.; Li, F.; Bai, H.; Cong, R.; Zhang, C.; Lin, C.; Liu, M.; and Zhao, Y. 2021.
\newblock Towards Fast and Accurate Real-World Depth Super-Resolution: Benchmark Dataset and Baseline.
\newblock In \emph{CVPR}, 9229--9238.

\bibitem[{Hirschmuller and Scharstein(2007)}]{hirschmuller2007evaluation}
Hirschmuller, H.; and Scharstein, D. 2007.
\newblock Evaluation of cost functions for stereo matching.
\newblock In \emph{CVPR}, 1--8.

\bibitem[{Hui, Loy, and Tang(2016)}]{hui2016depth}
Hui, T.-W.; Loy, C.~C.; and Tang, X. 2016.
\newblock Depth map super-resolution by deep multi-scale guidance.
\newblock In \emph{ECCV}, 353--369.

\bibitem[{Jiang et~al.(2021)Jiang, Dai, Wu, and Loy}]{jiang2021focal}
Jiang, L.; Dai, B.; Wu, W.; and Loy, C.~C. 2021.
\newblock Focal frequency loss for image reconstruction and synthesis.
\newblock In \emph{ICCV}, 13919--13929.

\bibitem[{Kim, Ponce, and Ham(2021)}]{kim2021deformable}
Kim, B.; Ponce, J.; and Ham, B. 2021.
\newblock Deformable kernel networks for joint image filtering.
\newblock \emph{International Journal of Computer Vision}, 129(2): 579--600.

\bibitem[{Kingma and Ba(2014)}]{Kingma2014Adam}
Kingma, D.~P.; and Ba, J. 2014.
\newblock Adam: A Method for Stochastic Optimization.
\newblock \emph{Computer Science}.

\bibitem[{Li et~al.(2016)Li, Huang, Ahuja, and Yang}]{li2016deep}
Li, Y.; Huang, J.-B.; Ahuja, N.; and Yang, M.-H. 2016.
\newblock Deep joint image filtering.
\newblock In \emph{ECCV}, 154--169.

\bibitem[{Li et~al.(2019)Li, Huang, Ahuja, and Yang}]{li2019joint}
Li, Y.; Huang, J.-B.; Ahuja, N.; and Yang, M.-H. 2019.
\newblock Joint image filtering with deep convolutional networks.
\newblock \emph{IEEE transactions on pattern analysis and machine intelligence}, 41(8): 1909--1923.

\bibitem[{Lin et~al.(2023)Lin, Li, Hsiao, Ho, and Kong}]{lin2023catch}
Lin, X.; Li, Y.; Hsiao, J.; Ho, C.; and Kong, Y. 2023.
\newblock Catch Missing Details: Image Reconstruction with Frequency Augmented Variational Autoencoder.
\newblock In \emph{CVPR}, 1736--1745.

\bibitem[{Lu, Ren, and Liu(2014)}]{lu2014depth}
Lu, S.; Ren, X.; and Liu, F. 2014.
\newblock Depth enhancement via low-rank matrix completion.
\newblock In \emph{CVPR}, 3390--3397.

\bibitem[{Ma et~al.(2020)Ma, Rao, Cheng, Chen, Lu, and Zhou}]{ma2020structure}
Ma, C.; Rao, Y.; Cheng, Y.; Chen, C.; Lu, J.; and Zhou, J. 2020.
\newblock Structure-preserving super resolution with gradient guidance.
\newblock In \emph{CVPR}, 7769--7778.

\bibitem[{Mao et~al.(2023)Mao, Liu, Liu, Li, Shen, and Wang}]{mao2023intriguing}
Mao, X.; Liu, Y.; Liu, F.; Li, Q.; Shen, W.; and Wang, Y. 2023.
\newblock Intriguing findings of frequency selection for image deblurring.
\newblock In \emph{AAAI}, 1905--1913.

\bibitem[{Metzger, Daudt, and Schindler(2023)}]{metzger2023guided}
Metzger, N.; Daudt, R.~C.; and Schindler, K. 2023.
\newblock Guided Depth Super-Resolution by Deep Anisotropic Diffusion.
\newblock In \emph{CVPR}, 18237--18246.

\bibitem[{Qiao et~al.(2023)Qiao, Ge, Zhang, Zhou, Tosi, Poggi, and Mattoccia}]{qiao2023depth}
Qiao, X.; Ge, C.; Zhang, Y.; Zhou, Y.; Tosi, F.; Poggi, M.; and Mattoccia, S. 2023.
\newblock Depth Super-Resolution from Explicit and Implicit High-Frequency Features.
\newblock \emph{arXiv preprint arXiv:2303.09307}.

\bibitem[{Scharstein and Pal(2007)}]{scharstein2007learning}
Scharstein, D.; and Pal, C. 2007.
\newblock Learning conditional random fields for stereo.
\newblock In \emph{CVPR}, 1--8.

\bibitem[{Shi, Ye, and Du(2022)}]{shi2022symmetric}
Shi, W.; Ye, M.; and Du, B. 2022.
\newblock Symmetric Uncertainty-Aware Feature Transmission for Depth Super-Resolution.
\newblock In \emph{ACM MM}, 3867–3876.

\bibitem[{Silberman et~al.(2012)Silberman, Hoiem, Kohli, and Fergus}]{silberman2012indoor}
Silberman, N.; Hoiem, D.; Kohli, P.; and Fergus, R. 2012.
\newblock Indoor segmentation and support inference from rgbd images.
\newblock In \emph{ECCV}, 746--760.

\bibitem[{Song et~al.(2020)Song, Dai, Zhou, Liu, Li, Li, and Yang}]{song2020channel}
Song, X.; Dai, Y.; Zhou, D.; Liu, L.; Li, W.; Li, H.; and Yang, R. 2020.
\newblock Channel attention based iterative residual learning for depth map super-resolution.
\newblock In \emph{CVPR}, 5631--5640.

\bibitem[{Su et~al.(2019)Su, Jampani, Sun, Gallo, Learned-Miller, and Kautz}]{su2019pixel}
Su, H.; Jampani, V.; Sun, D.; Gallo, O.; Learned-Miller, E.; and Kautz, J. 2019.
\newblock Pixel-adaptive convolutional neural networks.
\newblock In \emph{CVPR}, 11166--11175.

\bibitem[{Sun et~al.(2021)Sun, Ye, Li, Li, Wang, and Xu}]{sun2021learning}
Sun, B.; Ye, X.; Li, B.; Li, H.; Wang, Z.; and Xu, R. 2021.
\newblock Learning scene structure guidance via cross-task knowledge transfer for single depth super-resolution.
\newblock In \emph{CVPR}, 7792--7801.

\bibitem[{Sun, Xu, and Shum(2010)}]{sun2010gradient}
Sun, J.; Xu, Z.; and Shum, H.-Y. 2010.
\newblock Gradient profile prior and its applications in image super-resolution and enhancement.
\newblock \emph{IEEE Transactions on Image Processing}, 20(6): 1529--1542.

\bibitem[{Tang, Chen, and Zeng(2021)}]{tang2021joint}
Tang, J.; Chen, X.; and Zeng, G. 2021.
\newblock Joint implicit image function for guided depth super-resolution.
\newblock In \emph{ACM MM}, 4390--4399.

\bibitem[{Tang et~al.(2021)Tang, Cong, Sheng, He, Zhang, Zhao, and Kwong}]{tang2021bridgenet}
Tang, Q.; Cong, R.; Sheng, R.; He, L.; Zhang, D.; Zhao, Y.; and Kwong, S. 2021.
\newblock Bridgenet: A joint learning network of depth map super-resolution and monocular depth estimation.
\newblock In \emph{ACM MM}, 2148--2157.

\bibitem[{Woo et~al.(2018)Woo, Park, Lee, and Kweon}]{woo2018cbam}
Woo, S.; Park, J.; Lee, J.-Y.; and Kweon, I.~S. 2018.
\newblock Cbam: Convolutional block attention module.
\newblock In \emph{ECCV}, 3--19.

\bibitem[{Xiong et~al.(2021)Xiong, Hsiang, He, Zhan, and Wu}]{xiong2021augmented}
Xiong, J.; Hsiang, E.-L.; He, Z.; Zhan, T.; and Wu, S.-T. 2021.
\newblock Augmented reality and virtual reality displays: emerging technologies and future perspectives.
\newblock \emph{Light: Science \& Applications}, 10(1): 216.

\bibitem[{Yan et~al.(2022{\natexlab{a}})Yan, Wang, Li, Zhang, Li, Li, and Yang}]{yan2022learning}
Yan, Z.; Wang, K.; Li, X.; Zhang, Z.; Li, G.; Li, J.; and Yang, J. 2022{\natexlab{a}}.
\newblock Learning complementary correlations for depth super-resolution with incomplete data in real world.
\newblock \emph{IEEE transactions on neural networks and learning systems}.

\bibitem[{Yan et~al.(2022{\natexlab{b}})Yan, Wang, Li, Zhang, Li, and Yang}]{yan2022rignet}
Yan, Z.; Wang, K.; Li, X.; Zhang, Z.; Li, J.; and Yang, J. 2022{\natexlab{b}}.
\newblock RigNet: Repetitive image guided network for depth completion.
\newblock In \emph{ECCV}, 214--230. Springer.

\bibitem[{Yang et~al.(2022)Yang, Cao, Zhang, and Tao}]{yang2022codon}
Yang, Y.; Cao, Q.; Zhang, J.; and Tao, D. 2022.
\newblock CODON: on orchestrating cross-domain attentions for depth super-resolution.
\newblock \emph{International Journal of Computer Vision}, 130(2): 267--284.

\bibitem[{Yuan et~al.(2023{\natexlab{a}})Yuan, Jiang, Li, Qian, Li, and Yang}]{yuan2023recurrent}
Yuan, J.; Jiang, H.; Li, X.; Qian, J.; Li, J.; and Yang, J. 2023{\natexlab{a}}.
\newblock Recurrent Structure Attention Guidance for Depth Super-Resolution.
\newblock \emph{arXiv preprint arXiv:2301.13419}.

\bibitem[{Yuan et~al.(2023{\natexlab{b}})Yuan, Jiang, Li, Qian, Li, and Yang}]{yuan2023structure}
Yuan, J.; Jiang, H.; Li, X.; Qian, J.; Li, J.; and Yang, J. 2023{\natexlab{b}}.
\newblock Structure Flow-Guided Network for Real Depth Super-Resolution.
\newblock \emph{arXiv preprint arXiv:2301.13416}.

\bibitem[{Zhang et~al.(2018)Zhang, Li, Li, Wang, Zhong, and Fu}]{zhang2018image}
Zhang, Y.; Li, K.; Li, K.; Wang, L.; Zhong, B.; and Fu, Y. 2018.
\newblock Image super-resolution using very deep residual channel attention networks.
\newblock In \emph{ECCV}, 286--301.

\bibitem[{Zhao et~al.(2023)Zhao, Zhang, Gu, Tan, Xu, Zhang, Timofte, and Van~Gool}]{zhao2023spherical}
Zhao, Z.; Zhang, J.; Gu, X.; Tan, C.; Xu, S.; Zhang, Y.; Timofte, R.; and Van~Gool, L. 2023.
\newblock Spherical space feature decomposition for guided depth map super-resolution.
\newblock \emph{arXiv preprint arXiv:2303.08942}.

\bibitem[{Zhao et~al.(2022)Zhao, Zhang, Xu, Lin, and Pfister}]{zhao2022discrete}
Zhao, Z.; Zhang, J.; Xu, S.; Lin, Z.; and Pfister, H. 2022.
\newblock Discrete cosine transform network for guided depth map super-resolution.
\newblock In \emph{CVPR}, 5697--5707.

\bibitem[{Zhong et~al.(2021)Zhong, Liu, Jiang, Zhao, Chen, and Ji}]{zhong2021high}
Zhong, Z.; Liu, X.; Jiang, J.; Zhao, D.; Chen, Z.; and Ji, X. 2021.
\newblock High-resolution depth maps imaging via attention-based hierarchical multi-modal fusion.
\newblock \emph{IEEE Transactions on Image Processing}, 31: 648--663.

\bibitem[{Zhou et~al.(2022{\natexlab{a}})Zhou, Huang, Fang, Fu, and Liu}]{zhou2022pan}
Zhou, M.; Huang, J.; Fang, Y.; Fu, X.; and Liu, A. 2022{\natexlab{a}}.
\newblock Pan-sharpening with customized transformer and invertible neural network.
\newblock In \emph{AAAI}, volume~36, 3553--3561.

\bibitem[{Zhou et~al.(2022{\natexlab{b}})Zhou, Huang, Li, Yu, Yan, Zheng, and Zhao}]{zhou2022adaptively}
Zhou, M.; Huang, J.; Li, C.; Yu, H.; Yan, K.; Zheng, N.; and Zhao, F. 2022{\natexlab{b}}.
\newblock Adaptively learning low-high frequency information integration for pan-sharpening.
\newblock In \emph{ACM MM}, 3375--3384.

\bibitem[{Zhou et~al.(2022{\natexlab{c}})Zhou, Huang, Yan, Yu, Fu, Liu, Wei, and Zhao}]{zhou2022spatial}
Zhou, M.; Huang, J.; Yan, K.; Yu, H.; Fu, X.; Liu, A.; Wei, X.; and Zhao, F. 2022{\natexlab{c}}.
\newblock Spatial-frequency domain information integration for pan-sharpening.
\newblock In \emph{ECCV}, 274--291.

\bibitem[{Zhu et~al.(2015)Zhu, Zhang, Bonev, and Yuille}]{zhu2015modeling}
Zhu, Y.; Zhang, Y.; Bonev, B.; and Yuille, A.~L. 2015.
\newblock Modeling deformable gradient compositions for single-image super-resolution.
\newblock In \emph{CVPR}, 5417--5425.

\end{thebibliography}

\end{document}